\documentclass[11pt,a4paper]{article}
\usepackage[hyperref]{emnlp2020}
\usepackage{times}
\usepackage{latexsym}

\usepackage{microtype}

\usepackage{graphicx}
\usepackage{amsmath}
\usepackage{amsfonts}
\usepackage{mathrsfs}
\usepackage{multirow}

\usepackage{amsthm}
\usepackage{booktabs}
\usepackage{algorithm}
\usepackage{algorithmic}
\usepackage{amssymb}

\aclfinalcopy 
\def\aclpaperid{2031} 

\title{Grid Tagging Scheme for Aspect-oriented Fine-grained\\ Opinion Extraction}

\author{
	Zhen Wu\textsuperscript{\normalfont 1 } \quad 
	Chengcan Ying\textsuperscript{\normalfont 1} \quad 
	Fei Zhao\textsuperscript{\normalfont 1} \quad 
	Zhifang Fan\textsuperscript{\normalfont 1} \quad 
	Xinyu Dai\textsuperscript{\normalfont 1 \thanks{$^\dagger$ Corresponding author.}} \quad 
	Rui Xia\textsuperscript{\normalfont 2} \\ 
	\textsuperscript{1}National Key Laboratory for Novel Software Technology, Nanjing University \\
	\textsuperscript{2}School of Computer Science and Engineering, Nanjing University of Science and Technology \\
	\texttt{\{wuz, yingcc, zhaof, fanzf\}@smail.nju.edu.cn} \\
	\texttt{daixinyu@nju.edu.cn, rxia@njust.edu.cn} 
}

\date{}

\begin{document}
\maketitle
\begin{abstract}
	Aspect-oriented Fine-grained Opinion Extraction (AFOE) aims at extracting aspect terms and opinion terms from review in the form of opinion pairs or additionally extracting sentiment polarity of aspect term to form opinion triplet. Because of containing several opinion factors, the complete AFOE task is usually divided into multiple subtasks and achieved in the pipeline. However, pipeline approaches easily suffer from error propagation and inconvenience in real-world scenarios. To this end, we propose a novel tagging scheme, Grid Tagging Scheme (GTS), to address the AFOE task in an end-to-end fashion only with one unified grid tagging task. Additionally, we design an effective inference strategy on GTS to exploit mutual indication between different opinion factors for more accurate extractions. To validate the feasibility and compatibility of GTS, we implement three different GTS models respectively based on CNN, BiLSTM, and BERT, and conduct experiments on the aspect-oriented opinion pair extraction and opinion triplet extraction datasets. Extensive experimental results indicate that GTS models outperform strong baselines significantly and achieve state-of-the-art performance.
	
\end{abstract}

\section{Introduction}
\begin{figure}[t]
	\centering
	\includegraphics[width=0.99\linewidth]{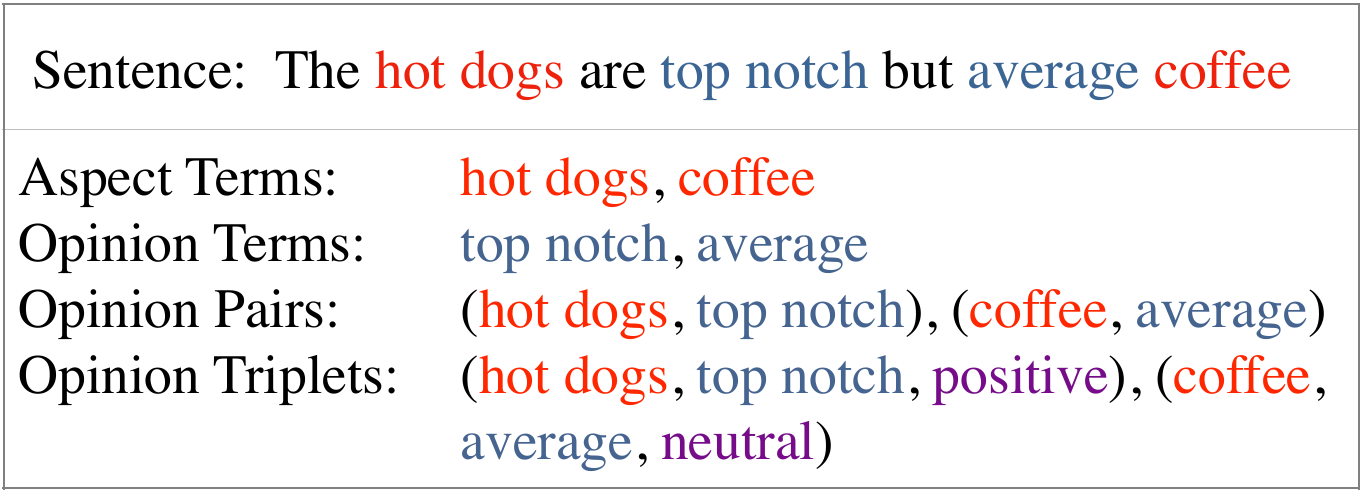}
	\caption{An example of aspect-oriented fine-grained opinion extraction. The spans highlighted in red are aspect terms. The terms in blue are opinion terms.}
	\label{opeexample}
\end{figure}

Aspect-oriented Fine-grained Opinion Extraction (AFOE) aims to automatically extract opinion pairs (\emph{aspect term}, \emph{opinion term}) or opinion triplets (\emph{aspect term}, \emph{opinion term}, \emph{sentiment}) from review text, which is an important task for fine-grained sentiment analysis~\cite{DBLP:journals/ftir/PangL07,DBLP:series/synthesis/2012Liu}. In this task, aspect term and opinion term are two key opinion factors. Aspect term, also known as opinion target, is the word or phrase in a sentence representing feature or entity of products or services. Opinion term refers to the term in a sentence used to express attitudes or opinions explicitly. For example, in the sentence of Figure~\ref{opeexample},  ``\emph{hot dogs}'' and ``\emph{coffee}'' are two aspect terms, ``\emph{top notch}'' and ``\emph{average}'' are two opinion terms.


To obtain the above two opinion factors, many works devote to the co-extraction of aspect term and opinion term in a joint framework~\cite{DBLP:conf/emnlp/WangPDX16,DBLP:conf/aaai/WangPDX17,DBLP:conf/emnlp/LiL17,DBLP:journals/taslp/YuJX19,DBLP:conf/acl/DaiS19}. However, the extracted results of these works are two separate sets of aspect term and opinion term, and they neglect the pair relation between them, which is crucial for downstream sentiment analysis tasks and has many potential applications, such as providing sentiment clues for aspect level sentiment classification~\cite{DBLP:conf/semeval/PontikiGPPAM14}, generating fine-grained opinion summarization~\cite{DBLP:conf/cikm/ZhuangJZ06} or analyzing in-depth opinions~\cite{DBLP:conf/emnlp/KobayashiIM07}, etc. 

Opinion pair extraction (OPE) is to extract all opinion pairs from a sentence in the form of (\emph{aspect term}, \emph{opinion term}). An opinion pair consists of an aspect term and a corresponding opinion term. This task needs to extract three opinion factors, i.e., aspect terms, opinion terms, and the pair relation between them. Figure 1 shows an example. We can see that the sentence ``\emph{the hot dogs are top notch and great coffee!}'' contains two opinion pairs, respectively (\emph{hot dogs}, \emph{top notch}) and (\emph{coffee}, \emph{average}) (the former is the aspect term, and latter represents the corresponding opinion term). OPE sometimes could be complicated because an aspect term may correspond to several opinion terms and vice versa. Despite the great importance of OPE, it is still under-investigated, and only a few early works mentioned or explored this task~\cite{DBLP:conf/kdd/HuL04,DBLP:conf/cikm/ZhuangJZ06,DBLP:conf/icdm/KlingerC13,DBLP:conf/acl/YangC13}.



By reviewing the aspect-based sentiment analysis (ABSA) ~\cite{DBLP:conf/semeval/PontikiGPPAM14} research, we can summarize two types of state-of-the-art pipeline approaches to extract opinion pairs: (I). Co-extraction~\cite{DBLP:conf/aaai/WangPDX17,DBLP:conf/acl/DaiS19}+Pair relation Detection (PD)~\cite{DBLP:conf/acl/XuLSY18}; (II). Aspect term Extraction (AE)~\cite{DBLP:conf/acl/XuLSY18}+Aspect-oriented Opinion Term Extraction (AOTE)~\cite{DBLP:conf/naacl/FanWDHC19}. Nevertheless, pipeline approaches easily suffer from error propagation and inconvenience in real-world scenarios. 

To address the above issues and facilitate the research of AFOE, we propose a novel tagging scheme, \textbf{G}rid \textbf{T}agging \textbf{S}cheme (GTS), which transforms opinion pair extraction into one unified grid tagging task. In this grid tagging task, we tag all word-pair relations and then decode all opinion pairs simultaneously with our proposed decoding method. Accordingly, GTS can extract all opinion factors of OPE in one step, instead of pipelines. Furthermore, different opinion factors are mutually dependent and indicative in the OPE task. For example, if we know ``\emph{average}'' is an opinion term in Figure~\ref{opeexample}, then ``\emph{coffee}'' is probably deduced as
an aspect term because ``\emph{average}'' is its modifier. To exploit these potential bridges, we specially design an inference strategy in GTS to yield more accurate opinion pairs. In the experiments, we implement three GTS models, respectively, with CNN, LSTM, and BERT, to demonstrate the effectiveness and compatibility of GTS.


Besides OPE, we find that GTS is very easily extended to aspect-oriented Opinion Triplet Extraction (OTE), by replacing the pair relation detection of OPE with specific sentiment polarity detection. OTE, also called aspect sentiment triplet extraction (ASTE)~\cite{DBLP:journals/corr/abs-1911-01616}, is a new fine-grained sentiment analysis task and aims to extract all opinion triplets (\emph{aspect term}, \emph{opinion term}, \emph{sentiment}) from a sentence. To tackle the task, ~\newcite{DBLP:journals/corr/abs-1911-01616} propose a two-stage framework and still extract the pair (\emph{aspect term}, \emph{opinion term}) in pipeline, thus suffering from error propagation. In contrast, GTS can extract all opinion triplets simultaneously only with a unified grid tagging task.


The main contributions of this work can be summarized as follows:
\begin{itemize}
	\item We propose a novel tagging scheme, Grid Tagging Scheme (GTS). To the best of our knowledge, GTS is the first work to address the complete aspect-oriented fine-grained opinion extraction, including OPE and OTE, with one unified tagging task instead of pipelines. Besides, this new scheme is easily extended to other pair/triplet extraction tasks from text.
	
	\item For the potential mutual indications between different opinion factors, we design an effective inference strategy on GTS to exploit them for more accurate extractions.
	
	\item We implement three GTS neural models respectively with CNN, LSTM, and BERT, and conduct extensive experiments on both tasks of OPE and OTE to verify the compatibility and effectiveness of GTS.
\end{itemize}

The following sections are organized as follows. Section~\ref{gridtaggingscheme} presents our proposed Grid Tagging Scheme. In Section~\ref{validationmodels}, we introduce the models based on GTS and the inference strategy. Section~\ref{experiments} shows experiment results. Section~\ref{relatedwork} and Section~\ref{conclusions} are respectively related work and conclusions. Our code and data will be available at \url{https://github.com/NJUNLP/GTS}.

\section{Grid Tagging Scheme}
\label{gridtaggingscheme}
In this section, we first give the task definition of Opinion Pair Extraction (OPE) and Opinion Triplet Extraction (OTE), then explain how the two tasks are represented in Grid Tagging Scheme. Finally, we present how to decode opinion pairs or opinion triplets according to the tagging results in GTS.

\subsection{Task Definition}
We first introduce the definition of the OPE task. Given a sentence $s=\{w_1, w_2, \cdots, w_n\}$ consisting $n$ words, the goal of the OPE task is to extract a set of opinion pairs $\mathcal{P}=\{(a, o)_m\}_{m=1}^{|\mathcal{P}|}$ from the sentence $s$, where $(a, o)_m$ is an opinion pair in $s$. The notations $a$ and $o$ respectively denote an aspect term and an opinion term. They are two non-overlapped spans in $s$.

As for the OTE task, it additionaly extracts the corresponding sentiment polarity of each opinion pair $(a, o)$, i.e., extracting a set of opinion tripelts $\mathcal{T}=\{(a, o, c)_m\}_{m=1}^{|\mathcal{T}|}$ from the given sentence $s$, where $c$ denotes the sentiment polarity and $c\in\{\text{positive, neutral, negative}\}$.

\subsection{Grid Tagging}
\label{gridtagging}

\begin{figure}[h]
	\centering
	\includegraphics[width=1.0\linewidth]{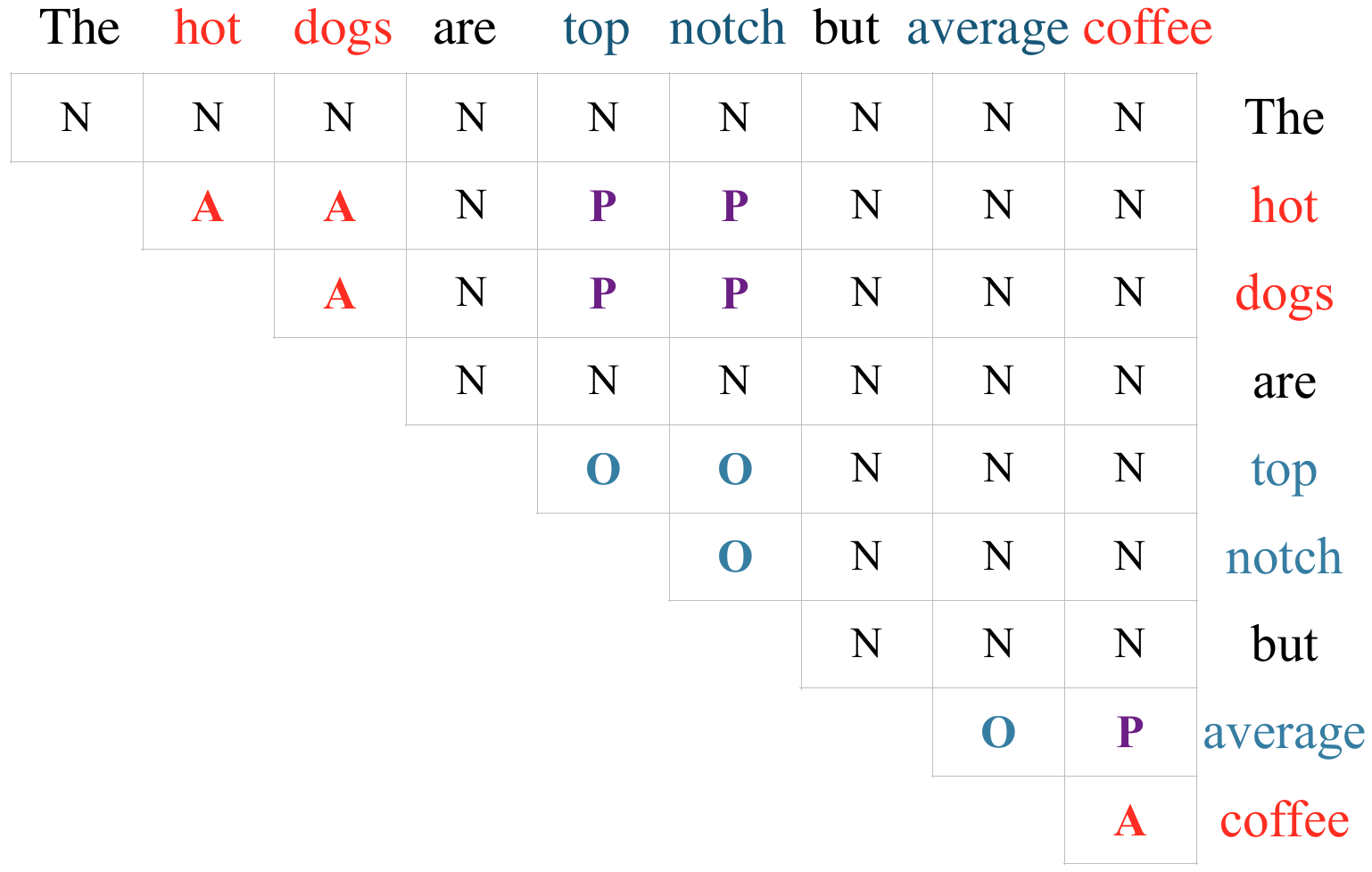}
	\caption{A tagging example with GTS for the OPE task. In the sentence, the spans highlighted in red are aspect terms and the spans in blue are opinion terms. }
	\label{opetagging}
\end{figure}

To tackle the OPE task, Grid Tagging Scheme (GTS) uses four tags \{\texttt{A}, \texttt{O}, \texttt{P}, \texttt{N}\} to represent the relation of any word-pair $(w_i, w_j)$ in a sentence. Here the word-pair $(w_i, w_j)$ is unordered and thus word-pair $(w_i, w_j)$ and $(w_j, w_i)$ have the same relation. The meanings of four tags can be seen in Table~\ref{opetags}. In GTS, the tagging result of a sentence is like a grid after displaying it in rows and columns. For simplicity, we adopt an upper triangular grid. Figure~\ref{opetagging} shows the tagging results of the sentence of Figure~\ref{opeexample} in GTS.

\begin{table}[!htbp]
	\resizebox{0.99\linewidth}{!}{
		\centering
		\begin{tabular}{c|c}
			\hline
			{Tags} & {Meanings} \\
			\hline
			\multirow{2}*{\texttt{A}}  & two words of word-pair $(w_i, w_j)$ belong to \\
			& the same aspect term.   \\
			\hline
			\multirow{2}*{\texttt{O}}   & two words of word-pair $(w_i, w_j)$ belong to \\
			& the same opinion term.\\
			\hline
			\multirow{3}*{\texttt{P}}   & two words of word-pair $(w_i, w_j)$ respectively \\
			&  belong to an aspect term and an opinion term,\\
			& and they form opinion pair relation.\\
			\hline
			\multirow{1}*{\texttt{N}}   & no above three relations for word-pair $(w_i, w_j).$\\
			\hline		
		\end{tabular}
	}
	\caption{The meanings of tags for the OPE task.}
	\label{opetags}
\end{table}

Specifically, the tag \texttt{A} represents that the two words of word-pair $(w_i, w_j)$ belong to the same aspect term. For example, the position of word-pair (\emph{hot}, \emph{dogs}) in Figure~\ref{opetagging} is the tag \texttt{A}. Similarly, the tag \texttt{O} indicates that the two words of word-pair $(w_i, w_j)$ exist in the same aspect term. Notably, GTS also considers the word-pair $(w_i, w_i)$, i.e., the relation of each word to itself, which can help represent a single-word aspect term or opinion term. The tag \texttt{P} represents that two words of word-pair $(w_i, w_j)$ respectively belong to an aspect term and an opinion term, and the two terms are an opinion pair, such as the word-pair (\emph{hot}, \emph{top}) and (\emph{dogs}, \emph{top}) in Figure~\ref{opetagging}. The last tag \texttt{N} denotes no relation between word-pair $(w_i, w_j)$.

\begin{figure}[h]
	\centering
	\includegraphics[width=1.0\linewidth]{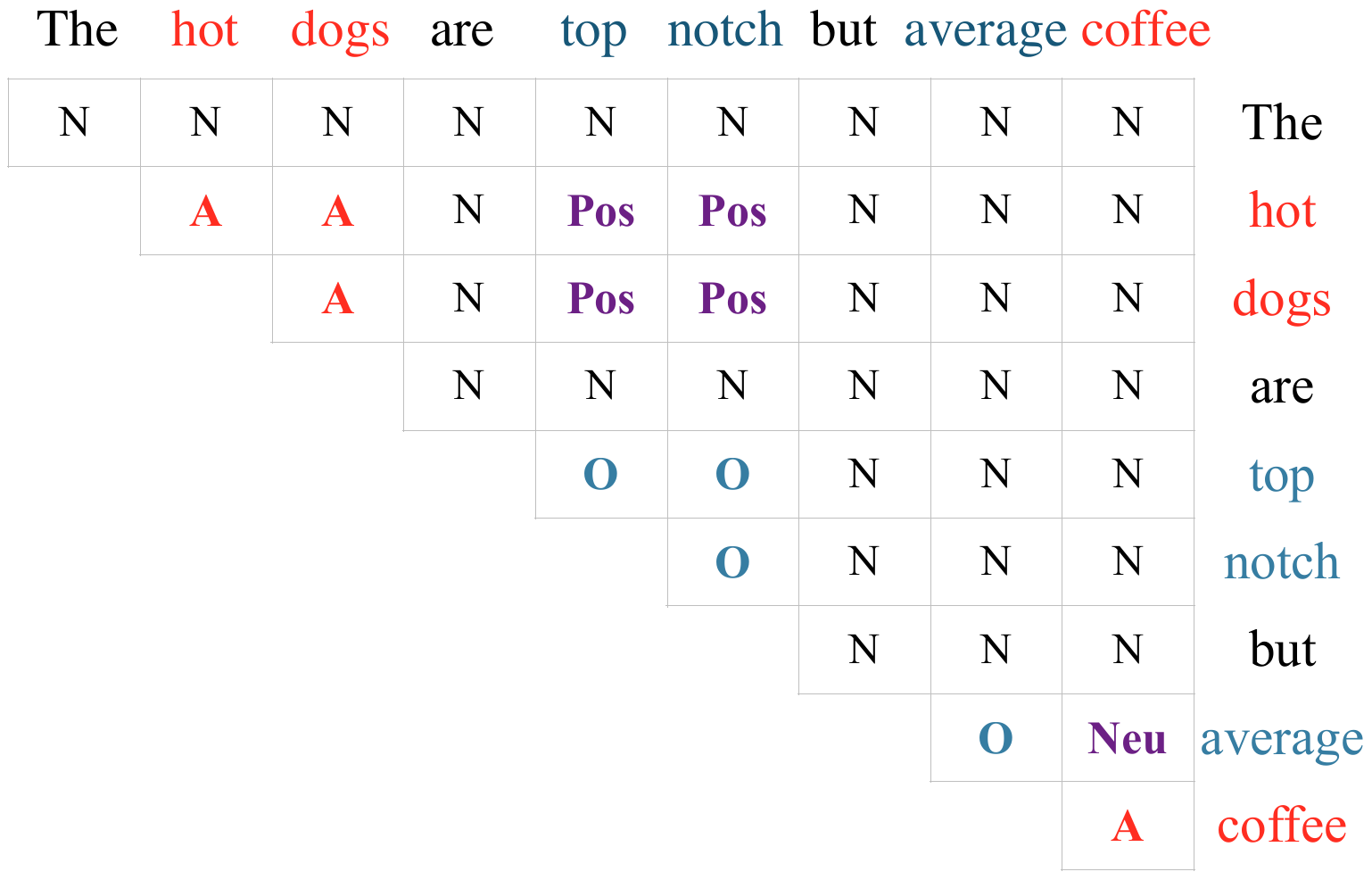}
	\caption{A tagging example for the OTE task.}
	\label{otetagging}
\end{figure}

To deal with the OTE task, GTS replaces the previous \texttt{P} tag with the specific sentiment label. To be specific, GTS adopts the tag set \{\texttt{A}, \texttt{O}, \texttt{Pos}, \texttt{Neu}, \texttt{Neg}, \texttt{N}\} to denote the relation of word-pair in the OTE task. The three tags \texttt{Pos}, \texttt{Neu}, \texttt{Neg} respectively indicate positive, neutral, or negative sentiment expressed in the opinion triplet consisting of the word-pair $(w_i, w_j)$. A tagging example of the OTE task is shown in Figure~\ref{otetagging}.

It can be concluded that Grid Tagging Scheme successfully transforms end-to-end aspect-oriented fine-grained opinion extraction into a unified tagging task by labeling the relations of all word-pairs.

\subsection{Decoding Algorithm}
In this subsection, we focus on how to decode the final opinion pairs or opinion triplets according to the tagging results of all word-pairs. In fact, various methods can be applied to obtaining these tagging results, and we adopt neural network models in this work (see Section~\ref{validationmodels}).

After obtaining the predicted tagging results of a sentence in GTS, we can extract opinion pairs or opinion triplets by strictly matching the relations of word-pairs as in Figure~\ref{opetagging} and Figure~\ref{otetagging}. However, it might get low recall due to abundant \texttt{N} tags in GTS. To address this issue, we relax matching constraints and design a simple but effective method to decode opinion pair or opinion triplet.

\begin{algorithm}[t]
	\small
	\caption{Decoding Algorithm for OPE}
	\label{alg:algorithm}
	\textbf{Input}: The tagging results $T$ of a sentence in GTS. $T(w_i, w_j)$ denotes the predicted tag of the word-pair $(w_i, w_j)$.\\
	\textbf{Output}: Opinion pair set $\mathcal{P}$ of the given sentence.
	\begin{algorithmic}[1] 
		\STATE Initialize the aspect term set $\mathcal{A}$, opinion term set $\mathcal{O}$, and opinion pair set $\mathcal{P}$ with $\varnothing$.
		\WHILE{a span left index $l\leq n$ and right index $r\leq n$}
		
		\IF {all $T(w_i, w_i) = \texttt{A} $ when $l\leq i \leq r$, meanwhile $T(w_{l-1}, w_{l-1}) \neq \texttt{A}$ and $T(w_{r+1}, w_{r+1}) \neq \texttt{A}$ }
		\STATE Regard the words $\{w_l,\cdots, w_r\}$ as an aspect term $a$, $\mathcal{A}\leftarrow \mathcal{A}  \cup  \{a\}$
		
		\ENDIF
		\IF {all $T(w_i, w_i) = \texttt{O}$ when $l\leq i \leq r$, meanwhile $T(w_{l-1}, w_{l-1}) \neq \texttt{O}$ and $T(w_{r+1}, w_{r+1}) \neq \texttt{O}$ }
		\STATE Regard the words $\{w_l,\cdots, w_r\}$ as an opinion term $o$, $\mathcal{O}\leftarrow \mathcal{O}  \cup  \{o\}$
		\ENDIF
		
		\ENDWHILE
		
		\WHILE{$a\in \mathcal{A}$ and $o\in \mathcal{O}$}
		
		\WHILE{$w_i\in a$ and $w_j\in o$}
		\IF {any $T(w_i, w_j) = \texttt{P}$}
		\STATE $\mathcal{P}\leftarrow \mathcal{P}  \cup  \{(a, o)\}$
		\ENDIF
		\ENDWHILE
		\ENDWHILE		
		\STATE \textbf{return} the set $\mathcal{P}$
	\end{algorithmic}
	\label{opedecoding}
\end{algorithm}

%
%
%
%
%

The decoding details for the OPE task are shown in Algorithm~\ref{opedecoding}. Firstly, we use the predicted tags of all $(w_i, w_i)$ word-pairs on the main diagonal to recognize aspect terms and opinion terms, without considering other word-pair constraints. As line 2 to line 9 of Algorithm~\ref{opedecoding} shows, the spans comprised of continuous \texttt{A} tags are regarded as aspect terms, and spans consisting of continuous \texttt{O} are detected as opinion terms. For an extracted aspect term $a$ and an opinion term $o$, we think they form an opinion pair on condition that at least one word-pair $(w_i, w_j)$ is labeled with the tag \texttt{P} when $w_i\in a$ and $w_j \in o$, as shown in line 11 to line 15.

For the OTE task, the decoding part is different from the OPE task from line 11 to line 15 of Algorithm~\ref{opedecoding}. Specifically, we count the predicted tags of all word-pairs $(w_i, w_j)$ when $w_i\in a$ and $w_j \in o$. The most predicted sentiment tag $c\in \{\texttt{Pos}, \texttt{Neu}, \texttt{Neg}\}$ is regarded as the sentiment polarity of the opinion triplet $(a, o, c)$. If their predicted tags do not belong to \{\texttt{Pos}, \texttt{Neu}, \texttt{Neg}\}, we think $a$ and $o$ cannot form an opinion triplet.

\section{Validation Models}
\label{validationmodels}
\begin{figure}[t]
	\centering
	\includegraphics[width=0.8\linewidth]{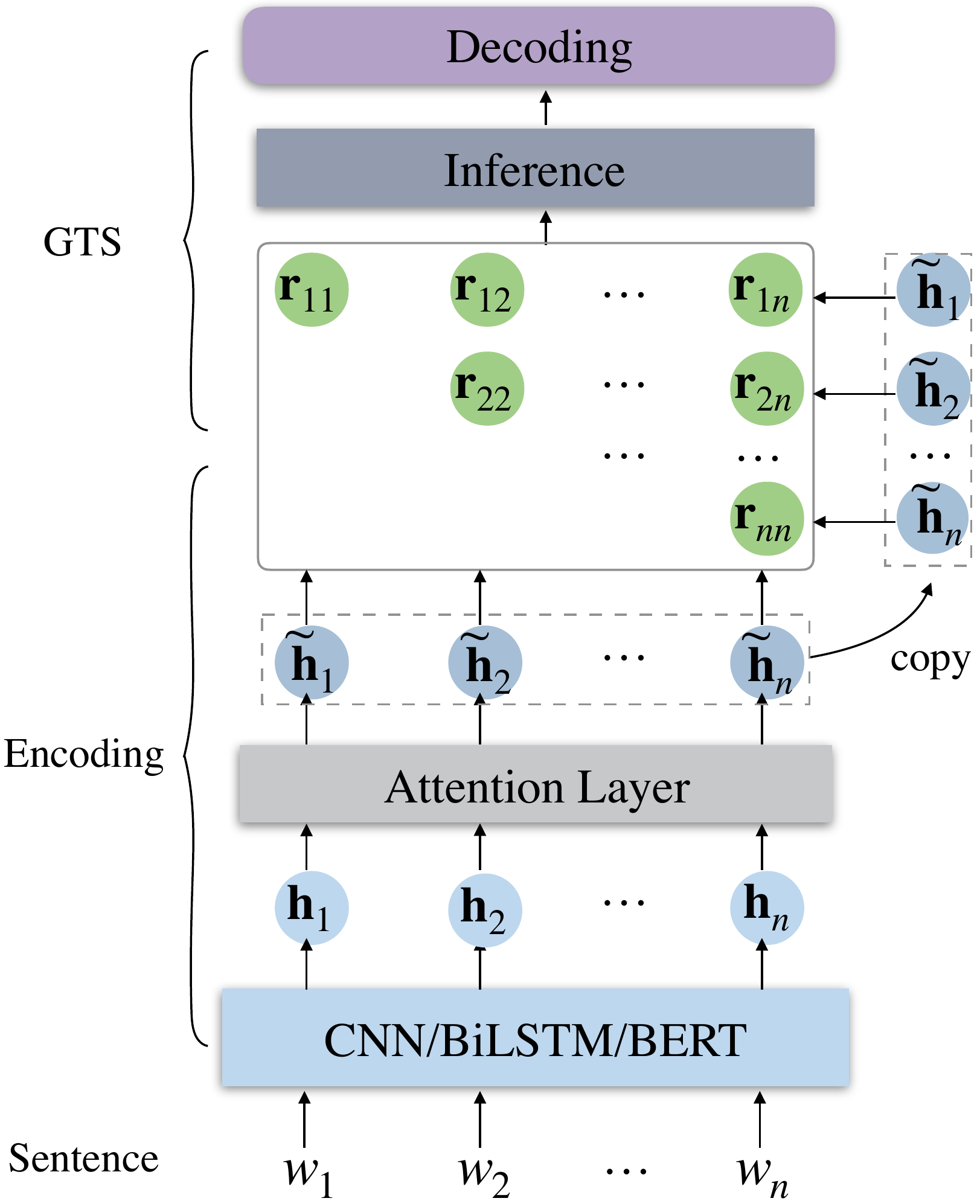}
	\caption{The overall architecture of neural models based on GTS.}
	\label{model}
\end{figure} 

To verify the effectiveness and good compatibility of GTS, we respectively tried three typical neural networks, i.e., CNN, LSTM, and BERT, as encoder implementations of GTS (Section~\ref{encoding}). Besides, different opinion factors in AFOE mutually rely on and can benefit each other. Therefore, we design an inference strategy to exploit these potential indications in Section~\ref{inferenceongts}. Figure~\ref{model} shows the overall architecture of GTS models.

\subsection{Encoding}
\label{encoding}
Given a sentence $s=\{w1, w_2, \cdots, w_n\}$, CNN, BiLSTM or BERT can be used as the encoder of GTS to generate the representation $\mathbf{r}_{ij}$ of the word-pair $(w_i, w_j)$.

\textbf{CNN}. We follow the design of state-of-the-art aspect term extraction model DE-CNN~\cite{DBLP:conf/acl/XuLSY18}. It employs 2 embedding layers and a stack of 4 CNN layers to encode the sentence $s$, then generates the feature representation $\mathbf{h}_i$ for each word $w_i$. Dropout~\cite{DBLP:journals/jmlr/SrivastavaHKSS14} is applied after the embedding and each ReLU activation. The details can be found in~\newcite{DBLP:conf/acl/XuLSY18}. 

\textbf{BiLSTM}. BiLSTM employs a standard forward Long Short-Term Memory (LSTM)~\cite{DBLP:journals/neco/HochreiterS97} and a backward LSTM to encode the sentence, then concatenate the hidden states in two LSTMs as the representation $\mathbf{h}_i$ of each word $w_i$.

\textbf{BERT}. BERT adopts subwords embedding, position embedding and segment embedding as the representation of subword, then employs a multi-layer bidirectional Transformer~\cite{DBLP:conf/nips/VaswaniSPUJGKP17} to generate the contextual represenations $\{\mathbf{h}_1, \mathbf{h}_2, \cdots, \mathbf{h}_n\}$ of the given sentence $s$. For a more comprehensive description, readers can refer to~\newcite{DBLP:conf/naacl/DevlinCLT19}.

To obtain a robust representation for word-pair $(w_i, w_j)$, we additionally employ an attention layer to enhance the connection between $w_i$ and $w_j$. The details are as follows:
\begin{align}
	u_{ij} &= \mathbf{v}^\top(\mathbf{W}_{a1}\mathbf{h}_i+ \mathbf{W}_{a2} \mathbf{h}_j+\mathbf{b}_a), \\
	\alpha_{ij} &= \frac{\exp(u_{ij})}{\sum_{k=1}^{n}\exp(u_{ik})}, \\
	\widetilde{\mathbf{h}_i} &= \mathbf{h}_i + \sum_{j=1}^{n}\alpha_{ij}\mathbf{h}_j,
\end{align}
where $\mathbf{W}_{a1}$ and $\mathbf{W}_{a2}$ are weight matrices, and $\mathbf{b}_a$ is the bias. Note that, the above attention is not applied on the representations of BERT, because BERT itself contains multiple self-attention layers.

Finally, we concatenate the enhanced representations of $w_i$ and $w_j$ to represent the word-pair $(w_i, w_j)$, i.e., $\mathbf{r}_{ij}=[\widetilde{\mathbf{h}_i}; \widetilde{\mathbf{h}}_j]$, where $[\cdot; \cdot]$ denotes the vector concatenation operation.

\subsection{Inference on GTS}
\label{inferenceongts}
As aforementioned, different opinion factors of AFOE are mutually indicative. Therefore, we design the inference strategy in GTS to exploit these potential indications for facilitating AFOE.

In Grid Tagging Scheme, let us consider what is helpful to detect the relation of word-pair $(w_i, w_j)$. First, relations between $w_i$ and other words (except $w_j$) can help detection. For example, if predicted tags of word-pairs consisting of $w_i$ contain $A$, the tag of word-pair $(w_i, w_j)$ is less possible to be $O$ and vice versa. So does the word $w_j$. Second, the previous prediction for $(w_i, w_j)$ heps infer the tag of $(w_i, w_j)$ of the current turn. To this end, we propose an inference strategy on GTS to exploit these indications by iterative prediction and inference. In the $t$-th turn, the feature representation $\mathbf{z}_{ij}^t$ and predicted probability distribution $\mathbf{p}_{ij}^t$ of word-pair $(w_i, w_j)$ can be calculated as follows:
\begin{align}
	\mathbf{p}_{i}^{t-1} &= \text{maxpooling}(\mathbf{p}_{i,:}^{t-1}), \label{prob_i}\\
	\mathbf{p}_{j}^{t-1} &= \text{maxpooling}(\mathbf{p}_{j,:}^{t-1}), \label{prob_j}\\
	\mathbf{q}_{ij}^{t-1} &= [\mathbf{z}_{ij}^{t-1}; \mathbf{p}_{i}^{t-1}; \mathbf{p}_{j}^{t-1}; \mathbf{p}_{ij}^{t-1}], \\
	\mathbf{z}_{ij}^t &= \mathbf{W}_{q}\mathbf{q}_{ij}^{t-1} +\mathbf{b}_{q}, \\
	\mathbf{p}_{ij}^t &= \text{softmax}(\mathbf{W}_{s}\mathbf{z}_{ij}^t +\mathbf{b}_{s}).
\end{align}

In the above process, $\mathbf{p}_{i,:}^{t-1}$ represents all predicted probability between the word $w_i$ and other words. In fact, $\mathbf{p}_{i,:}^{t-1} = (\mathbf{p}_{1:i,i}^{t-1}, \mathbf{p}_{i,i:n}^{t-1})$ in GTS as we use the upper triangular grid. Equation~\ref{prob_i} and ~\ref{prob_j} aim to help infer the possible tags for $(w_i, w_j)$ by observing predictions between $w_i$/$w_j$ and other words. The initial predicted probability $\mathbf{p}_{ij}^0$ and representation $\mathbf{z}_{ij}^0$ of $(w_i, w_j)$ is set as:
\begin{align}
	\mathbf{p}_{ij}^0 &= \text{softmax}(\mathbf{W}_{s}\mathbf{r}_{ij} +\mathbf{b}_{s}), \\
	\mathbf{z}_{ij}^0 &= \mathbf{r}_{ij}.
\end{align}

Finally, the prediction $\mathbf{p}_{ij}^L$ in the final turn is used to extract fine-grained opinions according to Algorithm~\ref{alg:algorithm}. The $L$ is a hyperparameter denoting the inference times.

\subsection{Training Loss}
We use $y_{ij}$ to represent the ground truth tag of the word-pair $(w_i, w_j)$. The unified training loss for AFOP is defined as the cross entropy loss between grouhd truth distribution and predicted tagging distribution $\mathbf{p}_{ij}^L$ of all word-pairs:
\begin{equation}
	\mathcal{L}=- \sum_{i=1}^{n}\sum_{j=i}^{n}\sum_{k\in C}\mathbb{I}(y_{ij}=k)\log(p_{i,j|k}^{L}),
\end{equation}
where $\mathbb{I}(\cdot)$ is the indicator function, and $C$ denotes the label set. In the OPE task, $C$ is $\{\texttt{A}, \texttt{O}, \texttt{P}, \texttt{N}\}$. For the OTE task, the set $C$ is $\{\texttt{A}, \texttt{O}, \texttt{Pos}, \texttt{Neu}, \texttt{Neg}, \texttt{N}\}$.

\section{Experiments}
\label{experiments}
\subsection{Datasets and Metrics}
\begin{table}[!htbp]
	\resizebox{0.99\linewidth}{!}{
		\centering
		\begin{tabular}{cc|ccccc}
			\hline
			\multicolumn{2}{c|}{Datasets} & {\#S} & {\#A} & {\#O} & {\#P} & {\#T}\\
			\hline
			\multirow{3}*{14res}  & Train & 1,259 & 2,064 & 2,098 & 2,356 & 2,356 \\
			& Dev & 315 & 487 & 506 & 580 & 580\\
			& Test & 493 & 851 & 866 & 1,008 & 1,008\\
			\hline
			\multirow{3}*{14lap}  & Train & 899 & 1,257 & 1,270 & 1,452 & 1,452\\
			& Dev & 225 &  332 & 313 & 383 & 383\\
			& Test & 332 &  467 & 478 & 547 & 547\\
			\hline
			\multirow{3}*{15res}  & Train & 603 & 871 & 966 & 1,038 & 1,038\\
			& Dev & 151 & 205 & 226 & 239 & 239\\
			& Test & 325 & 436 & 469 & 493 & 493\\
			\hline
			\multirow{3}*{16res}  & Train & 863 & 1,213 & 1,329 & 1,421 & 1,421\\
			& Dev & 216 &  298 & 331 & 348 & 348\\
			& Test & 328 &  456 & 485 & 525 & 525\\
			\hline		
		\end{tabular}
	}
	\caption{Statistics of aspect-oriented fine-grained opinion extraction datasets. Here ``\#S'', ``\#A'', ``\#O'', ``\#P'', and ``\#T'' respectively denote the numbers of sentence, aspect term, opinon term, opinion pair, and opinion triplet. The ``res'' and ``lap'' represent datasets from restaurant domain or laptop domain.}
	\label{datasets}
\end{table}

\begin{table*}[htbp]
	\resizebox{0.98\textwidth}{!}{
		\centering
		\begin{tabular}{c|c c c|c c c|c c c|c c c}
			\hline
			\multirow{2}{*}{Methods} & \multicolumn{3}{c|}{14res} & \multicolumn{3}{c|}{14lap} & \multicolumn{3}{c|}{15res} & \multicolumn{3}{c}{16res}\\ 
			\cline{2-13}
			&P& R& F1 &P& R& F1&P& R& F1&P& R& F1 \\
			\hline
			
			\multicolumn{13}{c}{\emph{Pipeline: Co-extraction+The Pair Relation Detection}}\\
			\hline
			CMLA+Dis-BiLSTM &  \textbf{77.21} & 52.14 & 62.24  & 59.47& 45.23& 51.17 & 64.86 & 44.33 & 52.47 & 66.29 & 50.82 & 57.33 \\
			CMLA+C-GCN & 72.22& 56.35& 63.17  & 60.69 & 47.25 & 53.03 & 64.31 & 49.41 & 55.76 & 66.61 & 59.23 & 62.70 \\
			RINANTE+C-GCN &71.07 & 59.45 & 64.69& \underline{67.38} & 52.10 & 58.76 & 65.52 & 42.74 & 51.73 & - & - & - \\
			\hline
			
			\multicolumn{13}{c}{\emph{Pipeline: Aspect Term Extraction+Aspect-oriented Opinion Term Extraction}}\\
			\hline			
			BiLSTM-ATT+Distance & 47.09&39.40 & 42.90& 38.85 & 29.20 & 33.34 & 39.63 & 33.95 & 36.57&43.60&39.65 & 41.53\\ 
			BiLSTM-ATT+Dependency & 56.31&48.93&52.36 & 31.58 & 28.84 & 30.15 & 58.26&42.19&48.94&64.48&48.85&55.59\\
			BiLSTM-ATT+IOG & 69.99&61.58&65.46 & 64.93 &44.56 &52.84 & 59.14 &56.38 &57.73 & 66.07&62.55&64.13\\
			DE-CNN+IOG &67.70&69.41&{68.55}& 59.59&{51.68}&{55.35} & 56.18&{60.08}&58.04 & 62.97&{66.22}&64.55\\ 
			RINANTE+IOG & 70.16 &65.47 &67.74& 61.76 &53.11 &57.10 & 63.24 & 55.57 &59.16 & -& -&-\\ 
			\hline
			
			\multicolumn{13}{c}{\emph{Our End-to-End GTS Models}}\\
			\hline
			GTS-CNN & {74.13}& \underline{69.49} & \underline{71.74}& \textbf{68.33}&\underline{55.04}&\underline{60.97}& \underline{66.81}&61.34&63.96 & \underline{70.48} & \underline{72.39} & \underline{71.42} \\		
			GTS-BiLSTM & 71.32&67.07&69.13& 61.53&54.31&57.69&  \textbf{67.76}&\underline{63.19}&\underline{65.39} & 70.32	&70.46&70.39\\		
			GTS-BERT & \underline{76.23}&\textbf{74.84}&\textbf{75.53}& 66.41&\textbf{64.95}&\textbf{65.67}&  66.40&\textbf{68.71}&\textbf{67.53} & \textbf{71.70} &\textbf{77.79}&\textbf{74.62} \\		
			\hline
	\end{tabular} }
	\caption{The experiment results on the OPE task (\%). Best and second-best results are respectively in bold and underline. The marker ``-'' represents that the original code of RINANTE method does not contain necessary resources for running on the dataset 16res.} 
	\label{operesults}
\end{table*}

To study aspect-oriented opinion term extraction,~\newcite{DBLP:conf/naacl/FanWDHC19} annotate and release four opinion pair datasets\footnote{https://github.com/NJUNLP/TOWE} based on SemEval Challenges~\cite{DBLP:conf/semeval/PontikiGPPAM14,DBLP:conf/semeval/PontikiGPMA15,DBLP:conf/semeval/PontikiGPAMAAZQ16}. However, they do not annotate the sentiment polarity of each opinion pair. The original SemEval Challenge datasets provide the annotation of aspect terms and the corresponding sentiment, while not the corresponding opinion terms. Thus we align the datasets of ~\newcite{DBLP:conf/naacl/FanWDHC19} and original SemEval Challenge datasets to build AFOE datasets. Table~\ref{datasets} shows their statistics, and we can observe that one sentence may contain multiple aspect terms or opinion terms. Besides, one aspect term may correspond to multiple opinion terms and vice versa.

To evaluate the performance of different methods, we use precision, recall, and F1-score as the evaluation metrics. The extracted aspect terms and opinion terms are regarded as correct only if predicted and ground truth spans are exactly matched.


\subsection{Experimental Settings}
Following the design of DE-CNN~\cite{DBLP:conf/acl/XuLSY18}, we use double embeddings to initialize the word vectors of GTS-CNN and GTS-BiLSTM, which contains a domain-general embedding from 300-dimension GloVe~\cite{DBLP:conf/emnlp/PenningtonSM14} pre-trained with 840 billion tokens and a 100-dimension domain-specific embedding trained with fastText~\cite{DBLP:journals/tacl/BojanowskiGJM17}. The CNN kernel size on domain-specific embedding is 3 and others are 5. In GTS-BiLSTM, the dimension of LSTM cell is set to 50. We adopt Adam optimizer~\cite{DBLP:journals/corr/KingmaB14} to optimize networks and the initial learning rate is 0.001. The dropout ~\cite{DBLP:journals/jmlr/SrivastavaHKSS14} is applied after embedding layer with probability 0.5. As for GTS-BERT, we use uncased $\text{BERT}_\text{BASE}$ version\footnote{https://github.com/google-research/bert} and set the learning rate to 5e-5. The mini-batch size is set to 32. The development set is used for early stopping. We run each model five times and report the average result of them.

\subsection{Results of Opinion Pair Extraction}

%
%

\begin{table*}[htbp]
	\resizebox{0.98\textwidth}{!}{
		\centering
		\begin{tabular}{c|c c c|c c c|c c c|c c c}
			\hline
			\multirow{2}{*}{Methods} & \multicolumn{3}{c|}{14res} & \multicolumn{3}{c|}{14lap} & \multicolumn{3}{c|}{15res} & \multicolumn{3}{c}{16res}\\ 
			\cline{2-13}
			&P& R& F1 &P& R& F1&P& R& F1&P& R& F1 \\
			\hline		
			Li-unified-R+$\text{PD}^{\S}$ & 41.44& \underline{68.79}&51.68& 42.25&42.78&42.47 & 43.34& 50.73 &46.69&38.19 &53.47 & 44.51\\ 
			Peng-unified-R+$\text{PD}^{\S}$  & 44.18 &62.99 &51.89 & 40.40 &47.24 &43.50 & 40.97 &\underline{54.68} &46.79&46.76 &62.97 & 53.62\\
			Peng-unified-R+IOG & 58.89& 60.41 &59.64 & 48.62&45.52 & 47.02 & 51.70 &46.04 &48.71 & 59.25&58.09&58.67\\
			IMN+IOG & 59.57 &63.88	&61.65 & 49.21&46.23&47.68  & 55.24&52.33&53.75 & -&-&-\\
			\hline
			
			GTS-CNN & \underline{70.79} &61.71 &\underline{65.94}& 55.93 &\underline{47.52} &\underline{51.38}& \underline{60.09} &53.57&\underline{56.64} & 62.63 &\textbf{66.98} &64.73 \\		
			GTS-BiLSTM & 67.28 &61.91 &64.49 & \textbf{59.42} &45.13 &51.30&  \textbf{63.26} &50.71 &56.29 & \underline{66.07} &65.05 &\underline{65.56}\\		
			GTS-BERT & \textbf{70.92}&\textbf{69.49}&\textbf{70.20}& \underline{57.52}&\textbf{51.92}&\textbf{54.58}&  59.29 &\textbf{58.07} &\textbf{58.67} & \textbf{68.58} &\underline{66.60} &\textbf{67.58} \\		
			\hline
	\end{tabular} }
	\caption{The experiment results on the OTE task (\%). Best and second-best results are respectively in bold and underline.  The results with ${\S}$ are retrieved from~\newcite{DBLP:journals/corr/abs-1911-01616}. The marker ``-'' represents that the original code of IMN method does not contain necessary resources for running on the dataset 16res.}
	\label{oteresults}
\end{table*}

\textbf{Compared Methods} We summarize the ABSA studies and combine the state-of-the-art methods as our strong OPE baselines. They include: (I). CMLA~\cite{DBLP:conf/aaai/WangPDX17} and RINANTE~\cite{DBLP:conf/acl/DaiS19} for the co-extraction of aspect term and opinion term (Co-extraction), Dis-BiLSTM and C-GCN~\cite{DBLP:conf/emnlp/Zhang0M18} for the Pair relation Detection (PD); (II). BiLSTM-ATT and DE-CNN~\cite{DBLP:conf/acl/XuLSY18} for Aspect term Extraction (AE), Distance~\cite{DBLP:conf/kdd/HuL04}, Dependency~\cite{DBLP:conf/cikm/ZhuangJZ06}, and IOG~\cite{DBLP:conf/naacl/FanWDHC19} for Aspect-oriented Opinion Term Extraction (AOTE). Note that, our GTS models do not use sentiment labels information when performing the OPE task. Table~\ref{operesults} shows the experiment results of different methods.

Observing two types of pipeline methods, we can find that the pipeline of AE+AOTE seems to perform better than Co-extraction+PD. Specifically, the method RINANTE+IOG outperforms RINANTE+C-GCN significantly on the datasets 14res and 15res, though C-GCN is a strong relation classification model. This indicates that the detection of opinion pair relation might be more difficult than aspect-oriented opinion term extraction. Besides, RINANTE+IOG also achieves better performances than another strong method DE-CNN+IOG respectively by the F1-score of 1.75\% and 1.12\% on the datasets 14lap and 15res, which validates the facilitation of co-extraction strategy for the aspect term extraction.

Compared with the strong pipelines DE-CNN+IOG and RINANTE+IOG, our three end-to-end GTS models all achieve obvious improvements, especially on the datasets 15res and 16res. Despite RINANTE using weak supervision to extend millions of training data, GTS-CNN and GTS-BiLSTM still obtain obvious improvements only through one unified tagging task without additional resources. This comparison shows that error propagations in pipeline methods limit the performance of OPE. There is no doubt that GTS-BERT achieves the best performance because of the powerful ability to model context. The results in Table~\ref{operesults} and above analysis consistently demonstrate the effectiveness of GTS for the OPE task.

\subsection{Results of Opinion Triplet Extraction}
\textbf{Compared Method} We use the latest OTE work proposed by \newcite{DBLP:journals/corr/abs-1911-01616} as the compared method. In addition, we also employ the state-of-the-art work IMN~\cite{DBLP:conf/acl/HeLND19} and the first step of ~\newcite{DBLP:journals/corr/abs-1911-01616} for extracting the (\emph{aspect term}, \emph{sentiment}) pair, then combine them with IOG as strong baselines. The experiment results are shown in Table~\ref{oteresults}.

We can observe that IMN+IOG outperforms Peng-unified-R+IOG obviously on the datasets 14res and 15res, because IMN uses multi-domain document-level sentiment classification data as auxiliary tasks. In contrast, GTS-CNN and GTS-BiLSTM still obtain about 3\% improvements in F1-score than IMN+IOG without requiring additional document-level sentiment data. The overall experiment results on the OTE task again validate the effectiveness of GTS. Furthermore, GTS-BERT outperforms GTS-CNN and GTS-BiLSTM only about 2\%-3\% on the datasets 15res and 16res, which to some extent shows the ability of the proposed tagging scheme itself besides BERT encoder.

\begin{table}[!htbp]
	\centering
	\resizebox{0.9\linewidth}{!}
	{
		
		\begin{tabular}{c|c c|c c}
			\hline
			\multirow{2}{*}{Methods} & \multicolumn{2}{c|}{14res} & \multicolumn{2}{c}{15res} \\ 
			\cline{2-5}
			&A& O &A& O \\
			
			\hline
			BiLSTM-ATT & 79.03 & 80.55 & 73.59 & 73.01 \\
			DE-CNN & \underline{81.90} & 80.57  & 75.24 & 73.07 \\
			CMLA & 81.22 & 80.48 & 76.03 & 74.67  \\
			RINANTE & 81.34 & \underline{83.33} & 73.38 & 75.40  \\
			\hline
			GTS-CNN & 81.82 & 83.07 & 77.33 & 75.23 \\
			GTS-BiLSTM & 81.10 & 82.62 &  \underline{78.44} & \underline{75.63} \\
			GTS-BERT & \textbf{83.82} & \textbf{85.04} &  \textbf{78.22} & \textbf{79.31}  \\
			\hline
		\end{tabular} 
	}
	\caption{The results of different methods on the extractions of aspect term and opinion term (\%) . The abbreviations ``A'' and ``O'' respectively denote the aspect term extraction and opinion term extraction.} 
	\label{aspectopinionextraction}
\end{table}

\subsection{Results of Aspects Term Extraction and Opinion Term Extraction}
To further analyze the performance of different methods, we also compare them on extractions of aspect term and opinion term. We only report F1-score of datasets 14res and 15res for limited space. The experiment results are shown in Tabel~\ref{aspectopinionextraction}.

Compared to GTS-CNN and GTS-BiLSTM, we can see that RINANTE achieves comparable or better results on the datasets 14res, while it performs worse on the OPE task. This comparison indicates that pipeline methods suffer from error propagation. According to the results on the dataset 15res, our GTS models not only can address the OPE task and OTE task in an end-to-end way, but also improve the performance of aspect term extraction and opinion term extraction. This is because our novel tagging scheme and inference strategy can exploit potential connections between different opinion factors to facilitate extraction.

%
%

\subsection{Ablation Study}

%
%

To investigate the effects of the attention mechanism and inference strategy on GTS models, we conduct ablation study on the OPE task. The experiment results are shown in Table~\ref{ablationstudy}.
\begin{table}[!htbp]
	\centering
	\resizebox{0.9\linewidth}{!}{
		\begin{tabular}{c|c|c|c|c}
			\hline
			
			\multirow{2}{*}{Methods} & 14res & 14lap & 15res & 16res\\ 
			\cline{2-5}
			& F1&F1& F1& F1 \\
			
			\hline
			GTS-CNN &{71.74}& {60.97}& 63.96 &  {71.42} \\		
			w/o attention & 70.33  &60.49 &63.09 & 70.88\\
			w/o inference & 68.92  &57.03 &61.81 & 66.66\\
			\hline
			
			GTS-BiLSTM &69.13&57.69&  {65.39} & 70.39\\		
			w/o attention &68.74&56.73 & 64.97 & 69.39\\
			w/o inference &67.55 & 55.94 &62.99 & 67.06\\
			\hline
			
		\end{tabular} 
	}
	\caption{Ablation study on the OPE task (\%) .} 
	\label{ablationstudy}
\end{table}

After removing the attention mechanism, the performance of the model GTS-CNN and GTS-BiLSTM drop slightly, which indicates that the attention mechanism enhances the connection between words. Comparing the full models with the versions w/o inference, we find that the former outperforms the latter significantly on all datasets. It is reasonable because the proposed inference strategy can leverage the potential bridges between different opinion factors and makes more comprehensive predictions. As for the model GTS-BERT w/o inference, it represents that the inference times is 0, and we show its results in the next section.

\subsection{Effects of Inference Times}

\begin{figure}[h]
	\centering
	\includegraphics[width=1.0\linewidth]{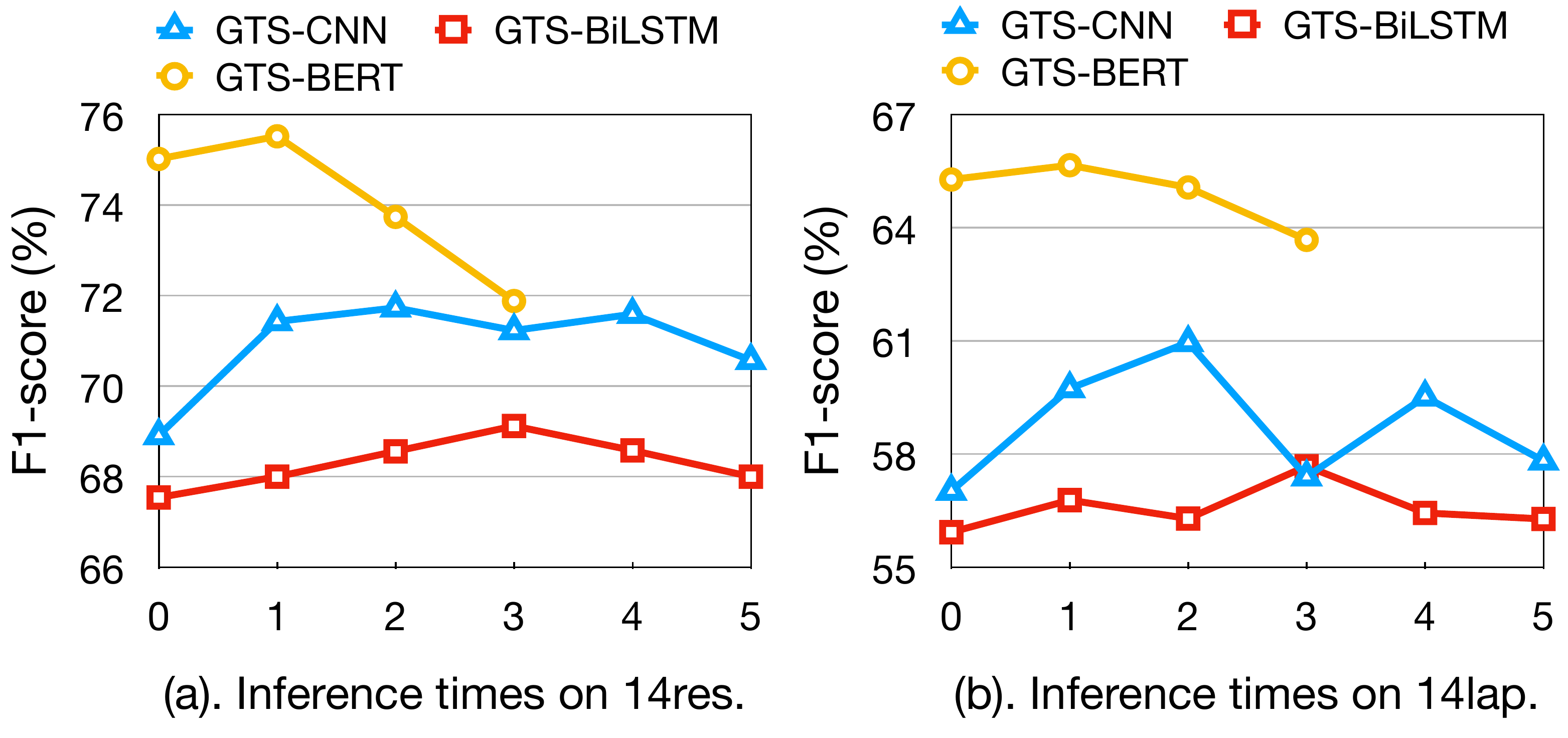}
	\caption{Effects of inference times on GTS models for the OPE task.}
	\label{inference times}
\end{figure} 

To investigate the effects of inference times on performance, we report the results of GTS models for the OPE task on the datasets 14res and 14lap with different inference times in Figure~\ref{inference times}.

It can be observed that the inference strategy brings significant improvements for the model GTS-CNN. On the whole, GTS-CNN and GTS-BiLSTM achieve the best results respectively with 2 and 3 inference times on two datasets, and GTS-CNN performs better than GTS-BiLSTM in different inference times. In contrast, GTS-BERT reaches a crest only with 1 time of inference because BERT has contained rich context semantics.


\section{Related Work}
\label{relatedwork}
In literature, only a few works mentioned or explored the opinion pair extraction.~\newcite{DBLP:conf/kdd/HuL04} employ frequent pattern mining to extract aspect terms, then regard the closest adjective to aspect term as the corresponding opinion term.~\newcite{DBLP:conf/cikm/ZhuangJZ06} adopt dependency-tree based templates to identify opinion pairs after extracting the aspect term set and opinion term set. Recently, some works adopt neural networks to perform the subtasks of OPE, such as co-extraction of aspect term and opinion term ~\cite{DBLP:conf/aaai/WangPDX17,DBLP:conf/acl/DaiS19}~\cite{DBLP:conf/acl/XuLSY18}. aspect term extraction~\cite{DBLP:conf/acl/XuLSY18}, and aspect-oriented opinion term extraction~\cite{DBLP:conf/naacl/FanWDHC19,DBLP:conf/aaai/WuZDHC20}, and finally combine them to accomplish OPE in pipeline. To avoid the error propagation of pipeline methods, some studies use joint learning based on traditional machine learning algorithms and hand-crafted features, including Imperatively Defined Factor graph (IDF)~\cite{klinger-cimiano-2013-bi}, joint inference based on IDF~\cite{DBLP:conf/icdm/KlingerC13}, and Integer Linear Programming (ILP)~\cite{DBLP:conf/acl/YangC13}. However, these methods heavily depend on the quality of hand-crafted features and sometimes perform worse than pipeline methods~\cite{DBLP:conf/icdm/KlingerC13}. 

The opinion triplet extraction is a new aspect-oriented fine-grained opinion extraction task~\cite{DBLP:journals/corr/abs-1911-01616}. Inspired by extracting (\emph{aspect term}, \emph{sentiment}) pair in a joint model~\cite{DBLP:conf/aaai/LiBLL19,DBLP:conf/acl/LuoLLZ19,DBLP:conf/acl/HeLND19}, ~\newcite{DBLP:journals/corr/abs-1911-01616} propose a two-stage framework to extract opinion triplets. In the first stage, they first use a neural model to extract the pair (\emph{aspect term}, \emph{sentiment}) and unpaired opinion terms, then detect the pair relation between aspect term and opinion terms in the second stage. We can see that the key opinon pair extraction of aspect term and opinion term is still accomplished in pipeline and their approach also suffers from error propagation.

\section{Conclusions}
\label{conclusions}
Aspect-oriented fine-grained opinion extraction (AFOE), including opinion pair extraction (OPE) and opinion triplet extraction (OTE), is usually achieved in the pipeline because of referring to multiple opinion factors, thereby suffering from error propagation. In this paper, we propose a novel scheme, Grid Tagging Scheme (GTS), to address this task in an end-to-end way. Through tagging the relations between all word-pairs, GTS successfully includes all opinion factors extraction of AFOE into a unified grid tagging task, and then uses the designed decoding algorithm to generate opinion pairs or opinion triplets. To exploit the potential mutual indications between different opinion factors, we design an effective inference strategy on GTS. Three different GTS models respectively based on CNN, BiLSTM, and BERT consistently indicate that our methods outperform strong baselines and achieve state-of-the-art performance on the opinion pair extraction and opinion triplet extraction. Further analysis also validates the effectiveness of GTS and the inference strategy.

\section*{Acknowledgments}
We would like to thank the anonymous reviewers for their valuable feedback. This work was supported by the NSFC (No. 61936012, 61976114) and National Key R\&D Program of China (No. 2018YFB1005102).


\bibliographystyle{acl_natbib}
\bibliography{emnlp2020}

\end{document}